\documentclass[journal]{IEEEtran}

\usepackage{hyperref}

\ifCLASSINFOpdf
   \usepackage[pdftex]{graphicx}
\else
\fi
\usepackage{subfig}

\usepackage{amsmath}
\usepackage{amssymb}
\begin{document}
%
\title{Beyond expectation: Deep joint mean and quantile regression for spatio-temporal problems}
%
%
%

\author{Filipe~Rodrigues,
        and~Francisco~C.~Pereira,~\IEEEmembership{Member,~IEEE}
\thanks{F.~Rodrigues is with the Technical University of Denmark (DTU), Bygning 116B, 2800 Kgs. Lyngby, Denmark. E-mail: rodr@dtu.dk}
\thanks{F.~C.~Pereira is with the Technical University of Denmark (DTU), Bygning 116B, 2800 Kgs. Lyngby, Denmark, and the Massachusetts Institute of Technology (MIT), 77 Massachusetts Avenue, 02139 Cambridge, MA, USA.}
\thanks{Manuscript received XXXX; revised XXXX.}}

%
%

\markboth{}%
{Rodrigues \MakeLowercase{\textit{et al.}}: Beyond expectation: Deep joint mean and quantile regression for spatio-temporal problems}
%



\maketitle

\begin{abstract}
Spatio-temporal problems are ubiquitous and of vital importance in many research fields. Despite the potential already demonstrated by deep learning methods in modeling spatio-temporal data, typical approaches tend to focus solely on conditional expectations of the output variables being modeled. In this paper, we propose a multi-output multi-quantile deep learning approach for jointly modeling several conditional quantiles together with the conditional expectation as a way to provide a more complete ``picture" of the predictive density in spatio-temporal problems. 

Using two large-scale datasets from the transportation domain, we empirically demonstrate that, by approaching the quantile regression problem from a multi-task learning perspective, it is possible to solve the embarrassing quantile crossings problem, while simultaneously significantly outperforming state-of-the-art quantile regression methods. Moreover, we show that jointly modeling the mean and several conditional quantiles not only provides a rich description about the predictive density that can capture heteroscedastic properties at a neglectable computational overhead, but also leads to improved predictions of the conditional expectation due to the extra information and a regularization effect induced by the added quantiles.
\end{abstract}


%
\IEEEpeerreviewmaketitle

\section{Introduction}

\IEEEPARstart{O}{ver} the last decade, deep learning has contributed to major advances in solving artificial intelligence problems in different domains such as speech recognition, visual object recognition, object detection and machine translation \cite{schmidhuber2015deep}. The ability of deep learning approaches to discover intricate structures in high dimensional data \cite{lecun2015deep} makes them very successful at tackling complex tasks such computer vision, perception, natural language understanding, etc. For similar reasons, and motivated by the contributions of deep learning to other domains, in recent years researchers have started to explore its applicability to model complex spatio-temporal data such as weather data \cite{xingjian2015convolutional}, urban mobility flows \cite{zhang2017deep}, crime data \cite{kang2017prediction} and video \cite{srivastava2015unsupervised}. This type of data is extremely ubiquitous and spans multiple research fields such as biology, economics, social sciences, transportation and environmental sensing. Therefore, developing models that are able to capture intricate spatial and temporal patterns and dependencies in the spatio-temporal data is of great importance to numerous research communities. 


Despite the early success already demonstrated by deep learning methods for approaching this type of problems in detriment of more traditional approaches based on probabilistic models, they often lack one fundamental characteristic: the ability to convey calibrated uncertainty estimates in their predictions. In our view, this results from an excessive focus of the research community on predicting conditional expectations of the form $\mathbb{E}[Y|X=x]$ as a function of $x$, and also from the lack of robust Bayesian inference methods that are able to scale to large neural networks. However, for many problems of interest, uncertainty estimates are of vital importance. Since the ultimate goal is to use the predictions of a deep neural network to make decisions, one needs to know how confident the model is when it produces a prediction, so that the decisions can be made accordingly. 
This is the case, for example, when assessing risk in financial applications, when predicting mobility demand for optimizing transportation systems, when forecasting energy consumption for market regulation, or when predicting product sales for managing stocks. 

In all the examples mentioned above, it is crucial to provide a more complete ``picture" of the forecasts that goes beyond the average relationship between inputs and target variables provided by standard deep learning approaches. To that end, we propose a deep learning approach for jointly modeling multiple conditional quantiles together with the conditional expectation. Since the different quantiles of a distribution are closely related, it is advantageous to approach the deep quantile regression problem from a multi-task learning perspective \cite{ruder2017overview}. As it turns out, doing so enforces consistency among the predicted quantiles, which makes it possible to address the embarrassing classical phenomenon of quantiles crossings, while also outperforming other state-of-the-art quantile regression approaches. Furthermore, as our empirical results demonstrate, adding conditional quantiles as additional outputs of a deep neural network introduces extra information about the data distribution and can have a regularization effect, which together can improve mean predictions. 

All the insights mentioned above are empirically verified by using a popular benchmark dataset for quantile regression and two large-scale spatio-temporal datasets from the transportation domain. Specifically, we consider the problems of taxi demand forecasting in NYC and traffic speeds forecasting in Copenhagen. In both these problems, it is essential to have fuller description of the predictive distribution that goes beyond the mean. For example, in taxi fleet coordination it is important to be prepared for potential disruptions caused by abnormal demand fluctuations. Similarly, when advising users about the best departure time so that they can confidently arrive on time at their destinations, it is important to account for possible delays and perhaps consider more pessimistic scenarios that can cause the travel to take longer than expected or connections to be missed. Quantile regression allows for this without assuming any particular parametric form for the target density, which can often be quite complex (\mbox{e.g.} skewed or multi-modal). Lastly, although this paper focus on the more difficult setting of spatio-temporal models with particular emphasis on transportation problems, it is important to note that the proposed approach extends trivially to simpler problems such as univariate time-series forecasting. 

In summary, the contributions of this paper are as follows:

\begin{itemize}
\item We propose a multi-output multi-quantile deep neural network architecture based on convolutional LSTM layers \cite{xingjian2015convolutional} for jointly predicting the mean and several quantiles of the predictive density at various spatial locations;
\item We employ the proposed deep learning approach for two fundamental transportation problems - taxi demand prediction and traffic speed forecasting - and we empirically show that the proposed approach leads to better quantiles than those produced by state-of-the-art quantiles regression methods and other popular methods for obtaining uncertainties from deep neural networks such as Monte Carlo dropout \cite{gal2016dropout};
\item We demonstrate how approaching the problem of multiple quantile estimation from a multi-task learning perspective can solve the classical quantile crossing phenomenon; 
\item Finally, we empirically show that adding the tilted loss for the quantiles (also know as the pinball loss) to the overall loss function adds relevant information about the target domain and can induce a regularization effect in the training of the neural network, which in turn can lead to better mean predictions. 
\end{itemize}

The remainder of this paper is organized as follows. In the next section, we review the relevant literature for this work. Section~\ref{sec:problem_formulation} formulates the problem of joint quantile regression in spatio-temporal data, while the proposed deep learning approach is introduced and explained in Section~\ref{sec:proposed_approach}. The corresponding experimental results are presented in Section~\ref{sec:experiments}. The paper ends with the conclusions (Section~\ref{sec:conclusion}).

\section{Literature Review}
\label{sec:lit_review}

For very reasonable practical and empirical reasons, the Gaussian distribution dominates the vast majority of practice and research in statistical models. To a lesser extent, but still overwhelmingly dominant, models tend to assume constant variance, i.e. homoscedasticity. In other words, that the error term, $\epsilon$ is independent of the independent variables of the model, $x$. Indeed, with a sufficiently complete specification and a balanced dataset, we should have all the possibly explainable variance covered in the model, leaving only the fundamental white noise to consider, i.e. $\epsilon \sim N(0, \sigma^2)$. Unfortunately, despite the increase in data, quality of sensors, and sophistication of models, in reality, it is often not possible to capture all involved variables in a phenomenon. 

In other words, there are often components of the distribution that we are predicting that relate to aspects we cannot capture, leading to non-constant variance (i.e. heteroscedasticity, manifested by $\epsilon \sim N(0, g(x))$), multi-modal distributions (e.g. mixtures), non-symmetrical or skewed distributions, long-tails, and so on. In many such cases, to overly focus on the mean may lead to frustrating results (e.g. the pooled mean of a Gaussian mixture has by definition low probability). 

On the other hand, for many models, there is sufficient research of their heteroscedastic variants and respective code available. Some notable examples include Gaussian processes \cite{Lazaro-gredilla11variationalheteroscedastic}, time series \cite{bollerslev1994arch}, neural networks \cite{wang2009nonlinear}, support vector regression \cite{hao2017pair} and of course many variations of linear regression models \cite{harvey1976estimating}. Still, a key limitation of most of these is the distribution form, typically a normal distribution. In other words, one models the non-constant variance for a predictive distribution that is necessarily Gaussian. This is usually done by adding the variance regression function $\sigma^2=g(x^+)$, where $x^+$ may correspond to additional data to the main model (e.g. information that there was an incident on the road may be more relevant to predict the variance than the mean of the travel time).

Differently, quantile regression provides us with a new perspective on the same problem: if one can predict any quantile of the target distribution, then why not directly estimate a prediction interval according to an arbitrary precision? For example, applying the functions for the 2.5\% and 97.5\% quantiles, one trivially determines the 95\% prediction interval. The function form of the quantile regression itself can be linear or in splines (Koenker \cite{Koenker05:quantile_regression_book}), non-linear, non-parametric with Gaussian processes \cite{antunes2017review,yang2018power} or vector-valued Reproducing Kernel Hilbert Space (RKHS) \cite{sangnier2016joint}. If we apply this principle for a significant number of different quantiles (e.g. every fifth quantile, 5, 10, 15 to 95), we can effectively obtain an approximation to the true predictive distribution that is much more robust to multi-modality or non-symmetry than assuming a parametric form. In fact, if we get a process that efficiently estimates quantiles to an arbitrarily high resolution, we can obtain accurate distributions to the precision we want. Of course, this doesn't come without a cost. A well-known challenge is the crossing quantiles - when a lower quantile function crosses a higher one. A substantial amount of work has addressed it (e.g. \cite{he1997quantile,schnabel2013simultaneous,chernozhukov2010quantile,bondell2010noncrossing}). A good summary is presented in \cite{sangnier2016joint}, who use constraints in a matrix-valued kernel to minimize the number of crossings. In fact, if assuming homoscedasticity on the individual quantile functions themselves, they guarantee the absence of crossings. In or work, Sangnier et al's RKHS model \cite{sangnier2016joint} is one of the baselines used for comparison. 

Another less mentioned issue is that, as we come close to the extremes (e.g. quantiles below 5\% and above 95\%), we face extremely sparse data. After all, these are, by many definitions, outliers. A consequence is that models tend to become much more unstable and hard to compare. Hence, despite the high relevance of wide intervals (e.g. 2.5\% to 97.5\%), literature typically focuses on tighter bounds, leaving the wider ones still as an open question. 

Alternative approaches exist for the specific estimation of prediction intervals. Mazloumi et al.~\cite{mazloumi2011prediction} provide a methodology for constructing prediction intervals for neural networks and quantifying the extent that each source of uncertainty contributes to total prediction uncertainty. The authors apply the methodology to bus travel time prediction and obtain quantitative decomposition of the prediction uncertainty into the effect of model structure and inputs data noise. Khosravi et al.~\cite{khosravi2011prediction} present two techniques, (i) delta, based on the interpretation of neural networks as nonlinear regressors, and (ii) Bayesian, for the construction of prediction intervals to account for uncertainties in travel time prediction. The results suggest that the delta technique outperforms the Bayesian technique in terms of narrowness of prediction intervals, while prediction intervals constructed with the Bayesian approach are more robust. Khosravi et al.~\cite{khosravi2011genetic} present a genetic algorithm-based method to automate the process of selecting the optimal neural network model specification. Model selection and parameter adjustments are performed through a minimization of a prediction interval--based cost function, with depends on the properties of the constructed prediction intervals. A review of other earlier neural network--based prediction interval methods can be found in~\cite{khosravi2011comprehensive}.

From the perspective of deep learning, emphasis has been given to modeling uncertainty in the model itself \cite{gal2016dropout}. In  \cite{gal2016dropout}, Gal and Ghahramani introduce the concept of Monte Carlo dropout as a method to approximate Bayesian inference in a deep neural network. Notice that this is conceptually different to what is proposed in this paper. In their case, the focus is the model itself (i.e. given priors on the weights and a model structure, what is their posterior distribution?), whereas in our case, we attempt to directly model the predictive distribution (i.e. given data available and a model structure, what is the predictive distribution of the target variable?). Nevertheless, the two approaches are obviously linked and nothing precludes us from using Monte Carlo dropout for approximating the predictive distribution and use that approximation to compute conditional quantiles. In fact, this approach is used in our experiments as one of the baselines.

\section{Problem formulation}
\label{sec:problem_formulation}

Suppose that we observe a dynamical system over a spatial region represented by an $M \times N$ grid. Let $y_{m,n,t}$ be the observed value in grid location $(m,n)$ at time $t$ of the variable that we wish to model, such that $\mathcal{Y} \in \mathbb{R}^{M \times N \times T}$ constitutes a spatio-temporal tensor corresponding to $M \times N$ time-series of duration $T$ at different spatial locations. In spatio-temporal forecasting, our goal is to build models that capture the spatial and temporal correlations present in the data and explore them to make forecasts $\{y_{m,n,t+k}, \forall (m,n) \in M \times N\}$ for a future time interval $t+k$ with $k \in \mathbb{N}$.

As previously mentioned, for most spatio-temporal problems of interest, it is essential to obtain a more complete representation of the predictive density to goes beyond the conditional expectation $\mathbb{E}[y_{m,n,t+k} | \mathcal{Y}_{1:t}]$, where we introduced the notation $\mathcal{Y}_{1:t}$ to denote all spatio-temporal information up to time $t$. For this purpose, we shall rely on conditional quantiles, since they yield crucial insights for many practical applications in which answers to important questions lie in modeling the tails of the conditional distribution. For $\tau \in (0,1)$, the conditional $\tau$-quantile for grid location $(m,n)$ at time $t$ is defined as the value $q_{m,n,t+k}^{(\tau)} \in \mathbb{R}: P(y_{m,n,t+k} < q_{m,n,t+k}^{(\tau)}) = \tau$. Given an lower and upper quantile, $q_{m,n,t+k}^{(\tau_l)}$ and $q_{m,n,t+k}^{(\tau_u)}$ respectively, it is possible to construct a prediction interval $I_{m,n,t+k}^{\alpha}$ consisting of a range of possible values within which the value $y_{m,n,t+k}$ is expected to lie with a given probability $\alpha = \tau_u - \tau_l$. Therefore, in many real-world problems, one is not only interested in estimating a single quantile function but rather several of them. 

Let $\{\tau_j\}_{j=1}^J$ be a set of $J$ quantile levels which we are interested in. A naive approach for estimating multiple quantiles would be to independently fit a function to each of them. However, this can be both computationally inefficient, since it implies fitting $J$ different functions, and lead to the violation of the basic principle that the quantiles should not cross, since the cumulative distribution function should be monotonically non-decreasing. With these problems in mind, in this paper, we propose approaching the problem of estimating the conditional expectation and conditional quantiles jointly by regarding them from a multi-task learning viewpoint \cite{ruder2017overview}. Since these estimation tasks can be highly related, then most of the relevant latent feature structures for modeling their spatio-temporal dynamics are expected to be similar. Our goal is then to, given a dataset $\mathcal{Y}_{1:t}$ of spatio-temporal observations up to time $t$, jointly predict the expected value $\mathbb{E}[y_{m,n,t+k} | \mathcal{Y}_{1:t}]$ and multiple conditional quantiles $\{q_{m,n,t+k}^{(\tau_j)}\}_{j=1}^J$, in order to accurately describe the predictive density at each grid cell $(m,n)$ at time $t+k$. As we shall empirically demonstrate, by approaching these estimation tasks jointly, it is possible to solve, at least to a great extent, the quantile crossings problems while reducing prediction error.

\section{Proposed approach}
\label{sec:proposed_approach}


In this section we describe the proposed multi-output multi-quantile deep neural network for jointly modeling the mean and multiple conditional quantiles in spatio-temporal problems. At the core of the proposed approach is the Convolutional LSTM layer \cite{xingjian2015convolutional}, or ConvLSTM for short. The ConvLSTM layer combines ideas from convolutional neural networks and long-short term memory (LSTM) cells to overcome some of the limitations of the latter for modeling spatio-temporal data. Although LSTMs have been very successfully in modeling sequence data \cite{schmidhuber2015deep,lecun2015deep}, they contain too much redundancy for spatial data. The ConvLSTM layer aims at addressing this shortcoming by introducing convolutional structures in both the input-to-state and state-to-state transitions of the LSTM. As a consequence, the inputs $\mathcal{Y}_1,\dots,\mathcal{Y}_t$, cell outputs $\mathcal{C}_1,\dots,\mathcal{C}_t$, hidden states $\mathcal{H}_1,\dots,\mathcal{H}_t$ and gates $i_t, f_t, o_t$ are now 3-dimensional tensors, whose two extra dimensions are spatial dimensions ($M \times N$). The key equations of a ConvLSTM can then be summarized as follows: 
\begin{align}
i_t &= \sigma(\mathcal{W}_{yi} * \mathcal{Y}_t + \mathcal{W}_{hi} * \mathcal{H}_{t-1}  + b_i), \nonumber\\
f_t &= \sigma(\mathcal{W}_{yf} * \mathcal{Y}_t + \mathcal{W}_{hf} * \mathcal{H}_{t-1} + b_f), \nonumber\\
\mathcal{C}_t &= f_t \odot \mathcal{C}_t + i_t \odot \mbox{tanh}(\mathcal{W}_{yc} * \mathcal{Y}_t + \mathcal{W}_{hc} * \mathcal{H}_{t-1} + b_c), \nonumber\\
o_t &= \sigma(\mathcal{W}_{yo} * \mathcal{Y}_t + \mathcal{W}_{ho} * \mathcal{C}_{t-1} + b_o), \nonumber\\
\mathcal{H}_t&= o_t \odot \mbox{tanh}(\mathcal{C}_t), 
\end{align}
where $\odot$ denotes the Hadamard product, and $*$ denotes the convolutional operator. Kindly notice that this is a simplified a version of the original ConvLSTM introduced in \cite{xingjian2015convolutional}, where the connections between the gates and the cell state have been omitted. 


\begin{figure}[!t]
\centering
\includegraphics[width=\linewidth]{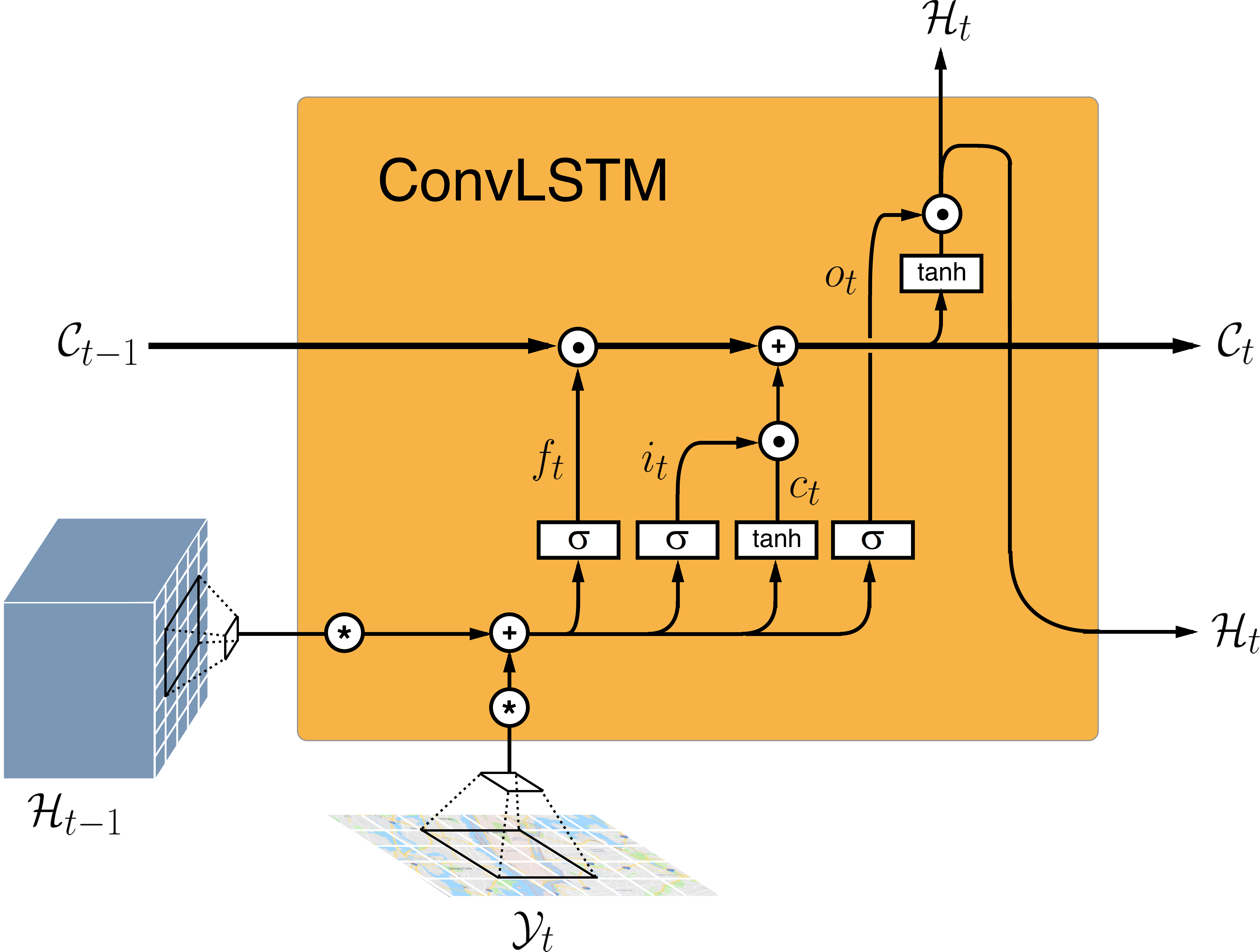}
\caption{Diagram of the architecture of the Convolutional LSTM layer, where $\odot$ denotes the Hadamard product, and $*$ denotes the convolutional operator.}
\label{fig:diagram_convlstm}
\end{figure}

By representing the cell state $\mathcal{C}_t$ and hidden state $\mathcal{H}_t$ as 3D tensors, such that $\mathcal{C}_t,\mathcal{H}_t \in \mathbb{R}^{M \times N \times S}$, and by introducing convolution operations, the ConvLSTM is able to determine the next state of a certain cell in the grid based on the inputs and previous states of its local neighbors, thereby providing it the ability to capture both temporal and spatial correlations in the data. This can be clearly observed from Figure~\ref{fig:diagram_convlstm}, which illustrates the inner operations of the ConvLSTM layer. 

\begin{figure}[!t]
\centering
\includegraphics[width=0.85\linewidth]{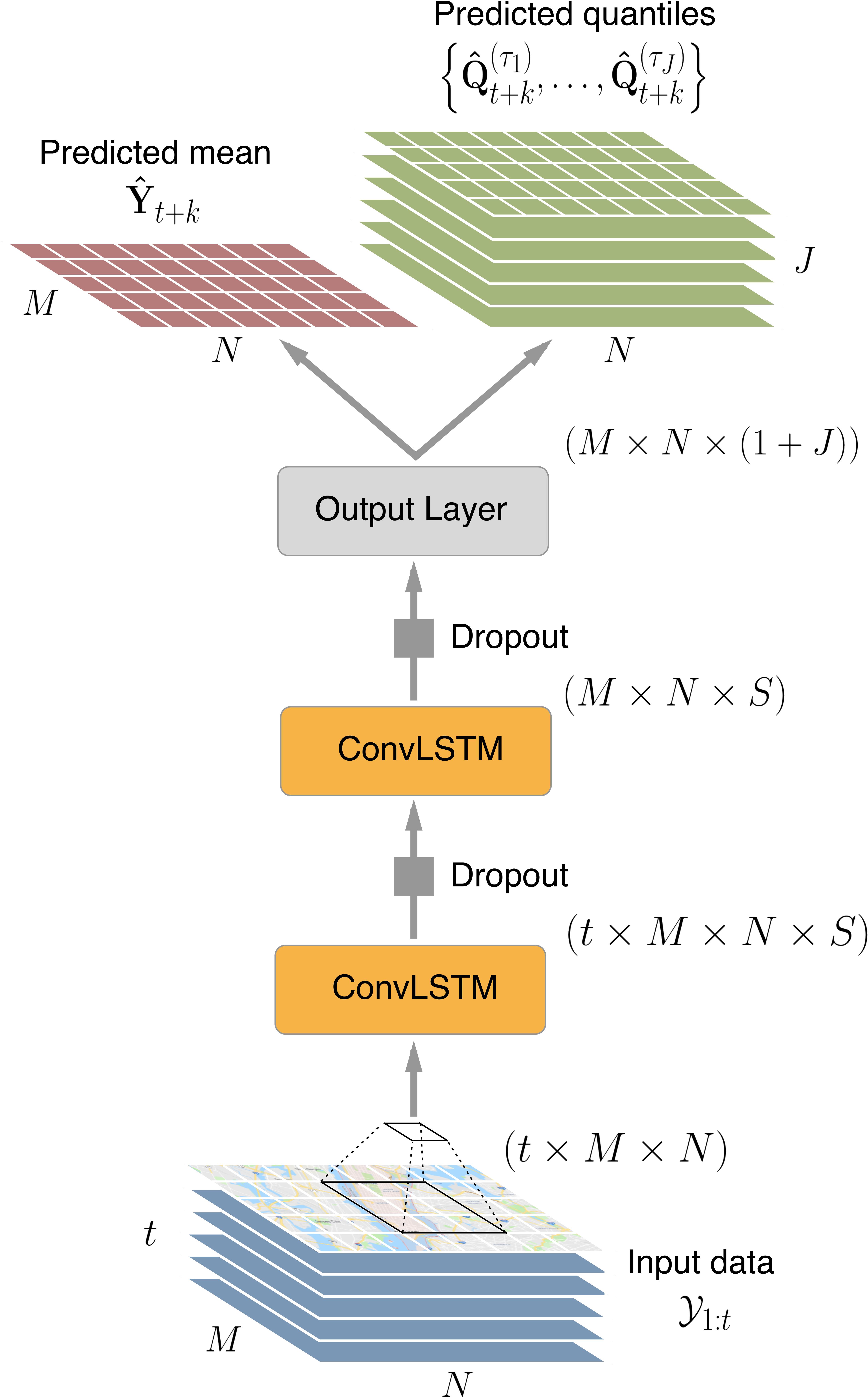}
\caption{Proposed neural network architecture.}
\label{fig:arch}
\end{figure}

The proposed architecture therefore consists of multiple convolutional LSTM layers stacked in order to capture an hierarchy of spatio-temporal patterns at various levels of abstraction as illustrated in Figure~\ref{fig:arch} for an example with two ConvLSTM layers. We can observe that the input of the neural network consists of a tensor $\mathcal{Y}_{1:t} \in \mathbb{R}^{t \times M \times N}$, although due practical reasons only a subset of the last $L$ observations is considered in practice. The raw input data is processed by the ConvLSTM layers in order to build a latent representation $\mathcal{H}_t \in \mathbb{R}^{M \times N \times S}$ in the hidden cell at time $t$ of the last ConvLSTM layer, encoding all the information required to describe the predictive distribution for each grid cell $(m,n)$ at time $t+k$. In practice, dropout \cite{srivastava2014dropout} and sometimes batch normalization \cite{ioffe2015batch} are used between the different layers in the proposed neural network architecture. 

\begin{figure}[!t]
\centering
\includegraphics[width=0.89\linewidth]{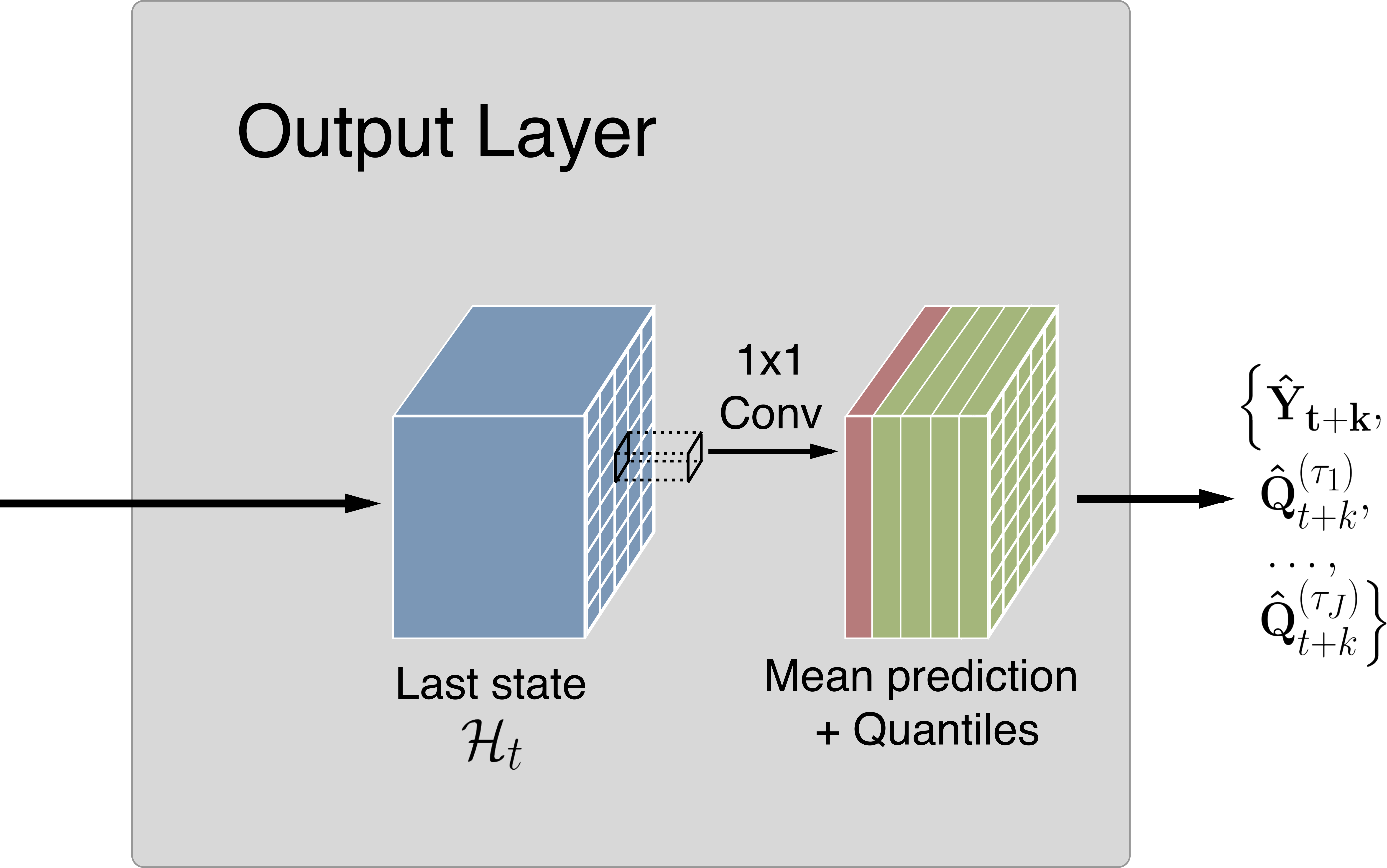}
\caption{Diagram of the architecture of the output layer.}
\label{fig:diagram_outputlayer}
\end{figure}

The goal of final stage of the proposed deep learning architecture is to turn this shared latent representation $\mathcal{H}_t$ into actual forecasts for the mean $\mathbb{E}[y_{m,n,t+k} | \mathcal{Y}_{1:t}]$ and multiple conditional quantiles $\{q_{m,n,t+k}^{(\tau_j)}\}_{j=1}^J$, without dramatically increasing the number of neural network parameters thus avoiding overfitting. For this purpose, we propose the use of an output layer that takes the latent state vector $\mathbf{h}_{m,n,t}$ corresponding to each spatial dimension $(m,n)$, and computes $1+J$ outputs - one for the predicted mean and $J$ for the multiple quantiles that we are interested in for obtaining a more complete description of the predictive distribution:
\begin{align}
\label{eq:output_layer}
\Big(\hat{y}_{m,n,t+k}, \hat{q}_{m,n,t+k}^{(\tau_1)}, \dots, \hat{q}_{m,n,t+k}^{(\tau_J)}\Big) = \mathbf{W}_{hp} \mathbf{h}_{m,n,t} + \mathbf{b}_p,
\end{align}
where $\mathbf{W}_{hp}$ and $\mathbf{b}_p$ are a $S \times (1+J)$ weight matrix and a $(1+J)$-dimensional bias vector respectively, that are shared among all grid cells (outputs), and whose sole purpose is to map latent representations learned into forecasts for the mean and quantiles. Alternately, this operation can be regarded as a $1 \times 1$ linear convolutional operation on the entire latent tensor representation $\mathcal{H}_t$:
\begin{align}
\Big(\mathbf{\hat{Y}}_{t+k}, \mathbf{\hat{Q}}_{t+k}^{\tau_1}, \dots, \mathbf{\hat{Q}}_{t+k}^{\tau_J}\Big) &= \mathbf{W}_{hp} * \mathcal{H}_t + \mathbf{b}_p
\end{align}
where $\mathbf{W}_{hp}$ and $\mathbf{b}_p$ are the weights and biases of the convolutional filter, and $\mathbf{\hat{Y}}_{t+k}, \mathbf{\hat{Q}}_{t+k}^{\tau_1}, \dots, \mathbf{\hat{Q}}_{t+k}^{\tau_J}$ are matrices containing the predictions for the mean and various quantiles for all grid cells. Figure~\ref{fig:diagram_outputlayer} illustrates this perspective. 

As the overall architecture depicted in Figure~\ref{fig:arch} shows, most of the latent structures are shared across different spatial locations and the predictors for the mean and different quantiles. If we regard these as related tasks, then the whole proposed architecture can be viewed from the perspective of multi-task learning as a hard-parameter sharing approach \cite{ruder2017overview}. As we shall see in Section~\ref{sec:experiments}, this imposes additional constraints among the different tasks, which can be very beneficial in practice. 

Having specified an appropriate neural network architecture that possesses the required properties, the next step is to figure out how to train it to produce the desired outputs: forecasts for the mean and conditional quantiles. In order to fit the mean at each time $t+k$, a loss function corresponding to the sum of the standard $\ell_2$-loss over all grid locations is used:
\begin{align}
\label{eq:mse_loss}
\ell_2(t+k) = \sum_{(m,n) \in M \times N} (y_{m,n,t+k} - \hat{y}_{m,n,t+k})^2.
\end{align}

\begin{figure}[!t]
\centering
\includegraphics[width=0.65\linewidth]{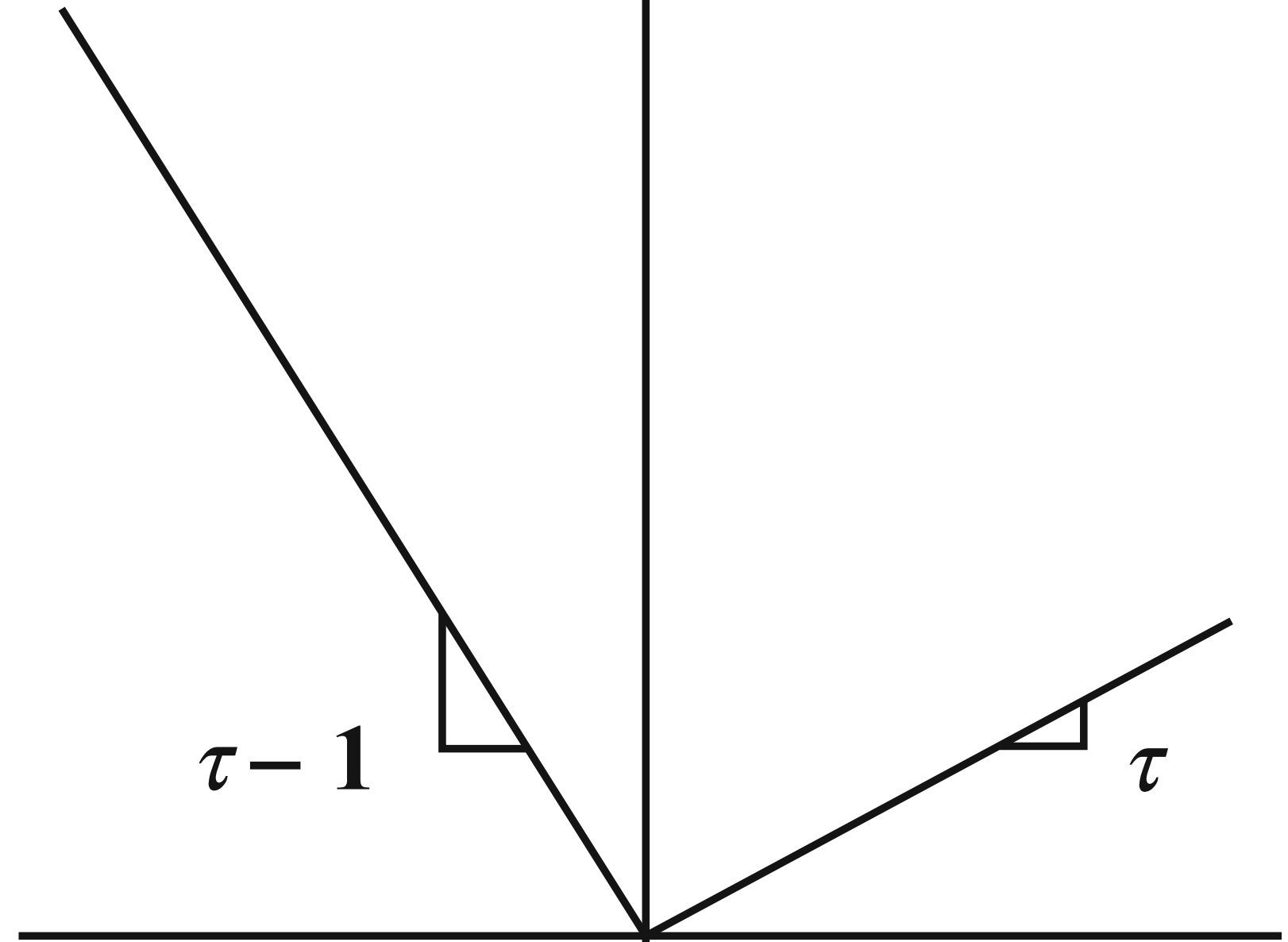}
\caption{Visual representation of the tilted loss function.}
\label{fig:tilted_loss}
\end{figure}

As for the conditional quantiles, fitting can be achieved by minimization of the tilted loss (also know as the pinball loss), defined as \cite{koenker_2005}
\begin{align}
\label{eq:tilted_loss}
\ell_\tau (r) = \begin{cases}
	\tau r, &\mbox{if } r \geq 0\\
	(\tau-1) r, &\mbox{if } r < 0\\
\end{cases},
\end{align}
or, more compactly as $\ell_\tau (r) = \max(\tau r, (\tau-1) r)$, where $r \in \mathbb{R}$ denotes the residual. 
The tilted loss results from the observation that the parameter $\mu$ that minimizes the $\ell_1$-loss, $\ell_1 = \sum_i |y_i - \mu|$, is an estimator of the unconditional median \cite{koenker1978regression}. The tilted loss can then be understood as a tilted version of the $\ell_1$-loss, where $\tau$ is a tilting parameter, as illustrated in Figure~\ref{fig:tilted_loss}. Thus, if an estimate falls above a given quantile (\mbox{e.g.} $\tau = 0.05$ quantile), the tilted loss $\ell_\tau$ is equal to the absolute value of the residual scaled by its probability $\tau$, thereby penalizing more overestimation than underestimation. As a result, one can obtain estimates of a conditional quantile $\tau$, by minimizing the empirical risk defined by $\ell_\tau$-loss. 

The proposed deep learning architecture can therefore be trained to produce estimates for the conditional mean and quantiles by minimizing the following objective function:
\begin{align}
\label{eq:objective_function}
E(t+k) &= \sum_{(m,n) \in M \times N} \bigg( \big(y_{m,n,t+k} - \hat{y}_{m,n,t+k}\big)^2 \nonumber\\
&+ \sum_{j=1}^J \max\Big(\tau_j \big(y_{m,n,t+k} - \hat{q}_{m,n,t+k}^{(\tau_j)}\big),\nonumber\\
& \big(\tau_j-1\big) \big(y_{m,n,t+k} - \hat{q}_{m,n,t+k}^{(\tau_j)}\big)\Big) \bigg).
\end{align}
In our implementation, this is done using automatic differentiation and the Adam optimizer \cite{kingma2014adam}. 

The objective function in (\ref{eq:objective_function}) can be regarded from two perspectives: (i) as a combination of multiple individual objective functions for the different outputs of the proposed neural network architecture that result from the multiple tasks, or (ii) as the objective function of a regression problem where the second term, resulting from the additional loss functions for the quantiles $\{\ell_{\tau_j}\}$, acts as a regularizer due to the hard-parameter sharing among the multiple tasks. As it turns out, the latter can have a significative impact in practice as a way of reducing overfitting and building more robust prediction models that also produce more accurate mean forecasts. 

A particularly important problem in conditional quantile estimation is quantile crossings \cite{he1997quantile}. In order to deal with this problem several approaches exist in the literature, such as stepwise estimation and simultaneous estimation with the use crossing constraints \cite{takeuchi2006nonparametric}. However, as pointed out by He \cite{he1997quantile}, the root of the problem to crossing is that quantile curves are computed individually so as to consistently estimate the conditional quantile functions in a broader class of models. Hence, although it would be possible to impose additional quantile crossing constraints to the our proposed objective function in (\ref{eq:objective_function}), we argue that it is possible to address the quantile crossing problem by limiting the flexibility of independent quantile regression neural network models via multi-task learning. In our proposed approach, this is done by having a common latent representation learned by the ConvLSTM layers for the multiple tasks, and by using hard-parameter sharing in the proposed output layer (see Eq.~\ref{eq:output_layer}). In the following section, we explore this perspective in detail and demonstrate empirically its effectiveness. 

\section{Experiments}
\label{sec:experiments}

The proposed multi-output multi-quantile deep learning approach for jointly modeling the conditional expectation together with several conditional quantiles, which we simply refer to as ``DeepJMQR" (as short for ``Deep Joint Mean and Quantile Regression"), was implemented in Keras \cite{chollet2015keras}. Source code is available in \url{http://fprodrigues.com/deep-jmqr/}. In this section, we perform an extensive set of experiments with different datasets, with emphasis on spatio-temporal data, for exploring different properties of the proposed approach and for providing a comparison with several state-of-the-art approaches.

\begin{figure*}[!t]
\centering
\includegraphics[width=0.96\linewidth]{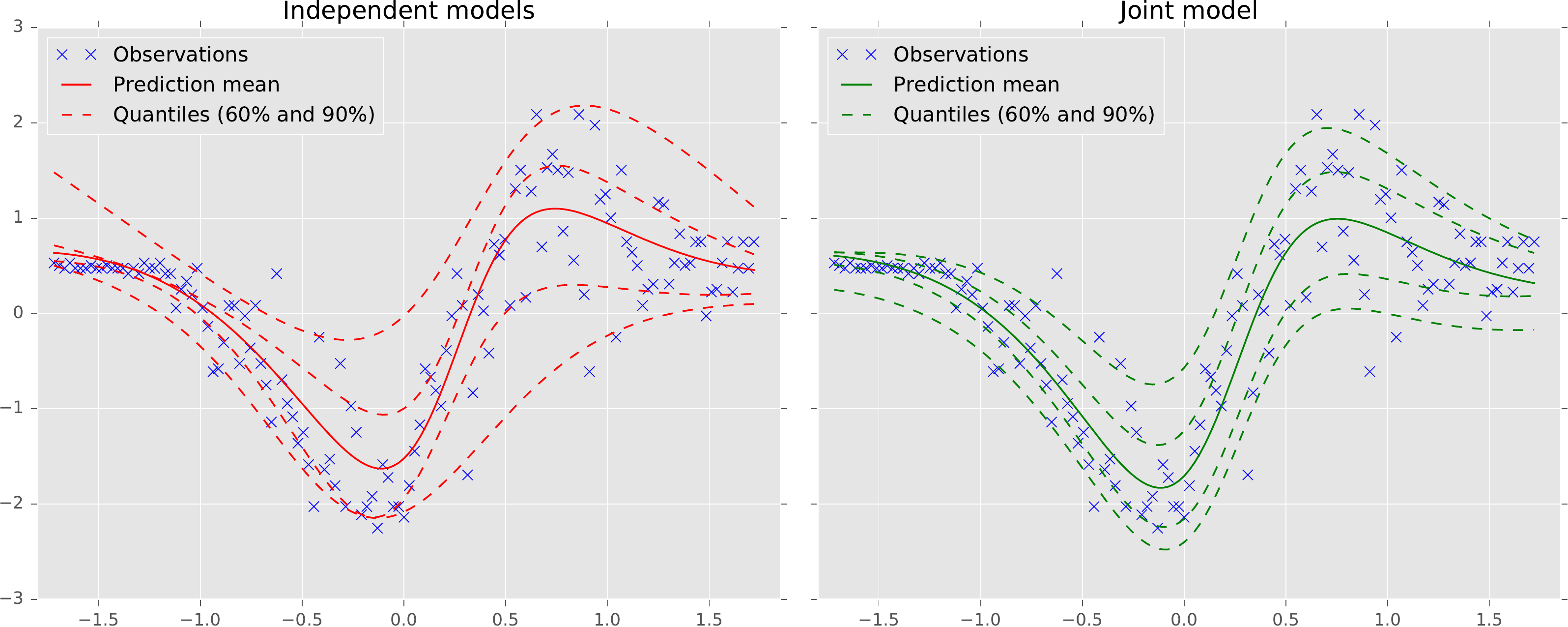}
\caption{Comparison between independent models and the proposed joint quantiles model for the Motorcycle dataset. The dashed lines correspond to the 5\%, 20\%, 80\% and 95\% quantiles, which allow the construction of 60\% and 90\% prediction intervals.}
\label{fig:motorcycle_comparison}
\end{figure*}

\begin{table*}[!t]
\renewcommand{\arraystretch}{1.1}
\caption{Results for Motorcycle dataset.}
\label{table:res_loss_motorcycle}
\centering
\begin{tabular}{l | l l | l l l}
Method & MAE & RMSE & Tilted Loss & Crossing Loss & Num. Crosses \\
\hline
Linear QR & 0.862 ($\pm$ 0.014) & 1.086 ($\pm$ 0.021) & 0.834 ($\pm$ 0.024) & 0.000 ($\pm$ 0.000) & 0.161 ($\pm$ 0.883) \\
Indep. DL & 0.420 ($\pm$ 0.015) & 0.520 ($\pm$ 0.010) &  0.420 ($\pm$ 0.016) & 0.004 ($\pm$ 0.002) & 2.194 ($\pm$ 1.090) \\
DeepJMQR & 0.413 ($\pm$ 0.013) & 0.515 ($\pm$ 0.009) &  0.398 ($\pm$ 0.014) & 0.000 ($\pm$ 0.000) & 0.742 ($\pm$ 0.506) \\
\end{tabular}
\end{table*}


\subsection{Motorcycle dataset}

We begin by performing experiments with a very popular heteroscedastic dataset - the motorcycle dataset from Silverman \cite{silverman1985some} - which is composed of 133 measurements of acceleration in the head of a crash test dummy vs. time in tests of motorcycle crashes. Many of the properties of this dataset, such as its size, shape and heteroscedastic noise, make them perfect for highlighting various characteristics of the proposed DeepJMQR neural network in comparison to the independent treatment of the multiple quantiles. 

Since this is a simpler problem than the spatio-temporal problems that will be considered later, we implemented a simpler neural network architecture that also allows us to demonstrate the generality of the proposed methodology. This neural network consists of 2 hidden layers: one with 50 hidden units with hyperbolic tangent activations, and another with 10 units with linear activations. The output layer also uses linear activations, but the number of units is either $1$ or $J+1$, depending whether we are considering independent models for the conditional expectation and various quantiles, or a joint model for everything. We refer to these as ``Indep. DL" and ``DeepJMQR", respectively. The independent model for the conditional mean is fit using the $\ell_2$-loss from (\ref{eq:mse_loss}), while the models for the quantiles are fit using the tilted loss $\ell_\tau$ (Eq.~\ref{eq:tilted_loss}).

In order to avoid visual clutter, we consider the problem of predicting the mean and 4 conditional quantiles: 5\%, 20\%, 80\% and 95\%, which allow us to construct 60\% and 90\% prediction intervals. The original dataset was split in 2/3 of the data points for training and 1/3 for evaluation. For the sake of comparison, we also fitted multiple standard linear quantile regression models independently. In order to evaluate the quality of the mean predictions, the mean absolute error (MAE) and root mean squared error (RMSE) were measured. As for evaluating the estimated quantiles, the sum of the tilted loss $\ell_\tau$ in the test set for all 4 quantiles was computed. Moreover, the multiple estimated quantiles were also evaluated in terms of the number of crosses and crossing loss (CL), defined as:
\begin{align}
\label{eq:cross_loss}
\mbox{CL} = \sum_{i=1}^{N_{\mbox{\scriptsize test}}} \sum_{j=1}^{J-1} \max\big(0, \hat{q}_{i}^{(\tau_j)} - \hat{q}_{i}^{(\tau_{j+1})}\big),
\end{align}
where it is assumed that $\tau_{j+1} > \tau_j, \forall j \in {1,\dots,J-1}$ and $N_{\mbox{\scriptsize test}}$ is used to denote the number test points. 

Due to the stochasticity introduced by the random initialization of the neural network weights, each experiment is repeated 30 times and the average results are reported, together with the corresponding standard deviations. Table~\ref{table:res_loss_motorcycle} shows the obtained results. We can observe that the deep-learning-based non-linear approaches (Indep. DL and DeepJMQR) significantly outperform their linear counterparts, by roughly halving the prediction error for the mean in terms of both MAE and RMSE, and by also halving the tilted loss in the test set. This is expectable, due to the added flexibility and complexity of the functional forms provided by neural networks when compared to linear methods (linear regression and linear quantile regression). However, we can observe that, in the case of the multiple independent models, the added complexity leads to an increase in the crossing loss. On the other hand, by considering the multiple quantile regression tasks jointly, we can verify that in the proposed DeepJMQR, the quantile crossing problem is significantly reduced. In our interpretation, this is a consequence of the multi-task learning approach, which enforces consistency and coherence among the different quantiles. The latter are very clear in Figure~\ref{fig:motorcycle_comparison}, which compares the estimated quantiles by multiple independent models and the ones predicted by the proposed DeepJMQR approach. Moreover, by also jointly modeling the mean together with the conditional quantiles, we can observe that the consistency constraints induced by DeepJMQR also lead to a reduction in MAE and RMSE when compared to using a single independent model for predicting the mean. 

\begin{figure}[!t]
\centering
\includegraphics[width=0.85\linewidth]{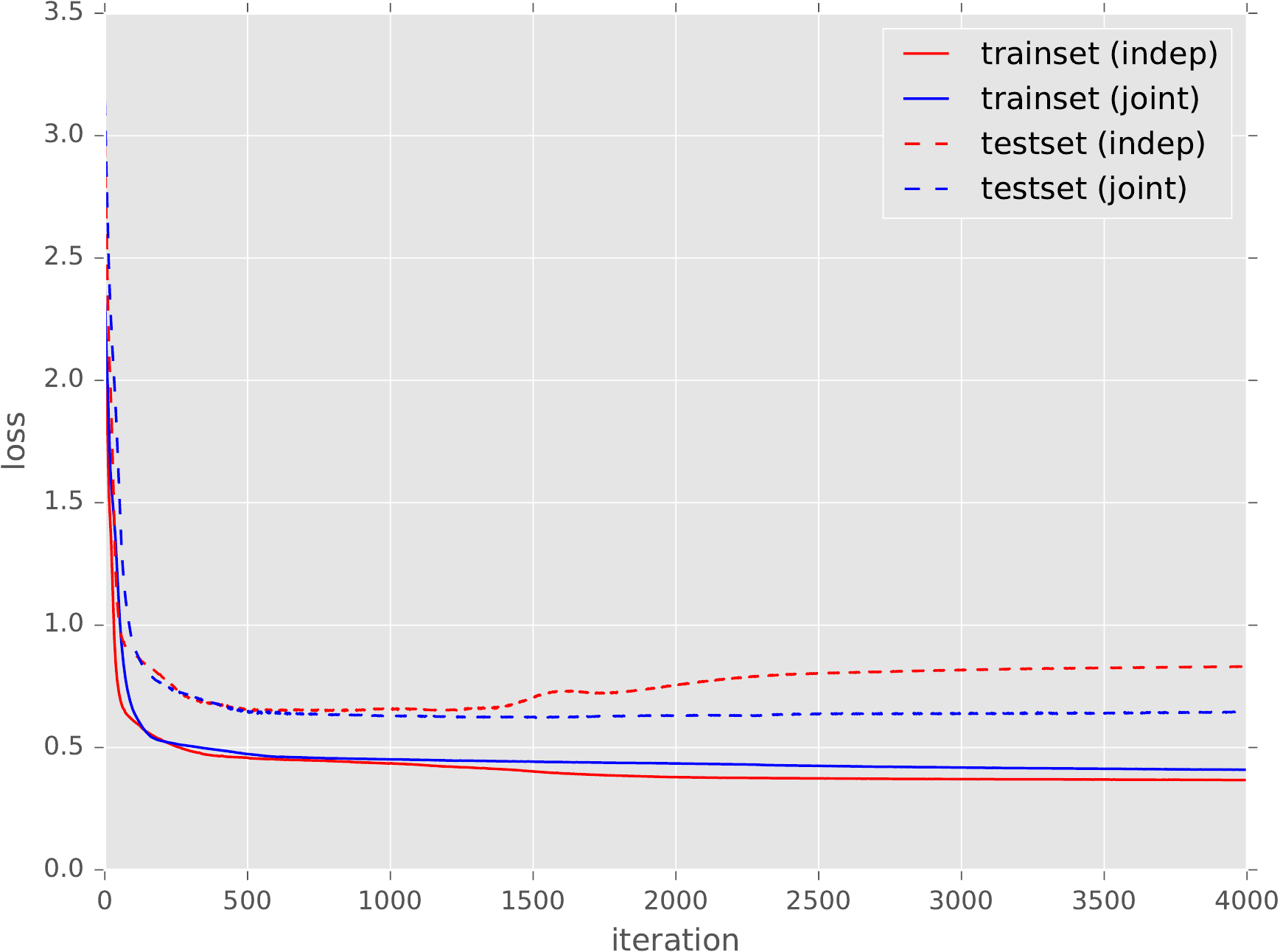}
\caption{Regularization effect of joint quantile learning.}
\label{fig:motorcycle_losses}
\end{figure}

Lastly, we increased the flexibility of the proposed neural network by replacing the linear activations of the second layer to hyperbolic tangets, and we let the training of the neural networks run for longer and studied the evolution of the objective function in both the train set and test set. Figure~\ref{fig:motorcycle_losses} shows this evolution for the proposed DeepJMQR and the independent models. For the latter, the losses of the individual models where summed up. Based on this figure, we can observe the regularization effect induced by the multi-task learning approach. We can verify that, at some point, the test loss of the independent models begins to increase while the train loss continues to decrease - a clear sign of overfitting. On the contrary, as the figure shows, the proposed DeepJMQR network is robust to overfitting, which we attribute to a regularization effect caused by it having to model a more complete description of the predictive distribution with the same underlying neural network structure. 

\subsection{Taxi demand in NYC}

\begin{figure}[!t]
\centering
\includegraphics[width=0.85\linewidth]{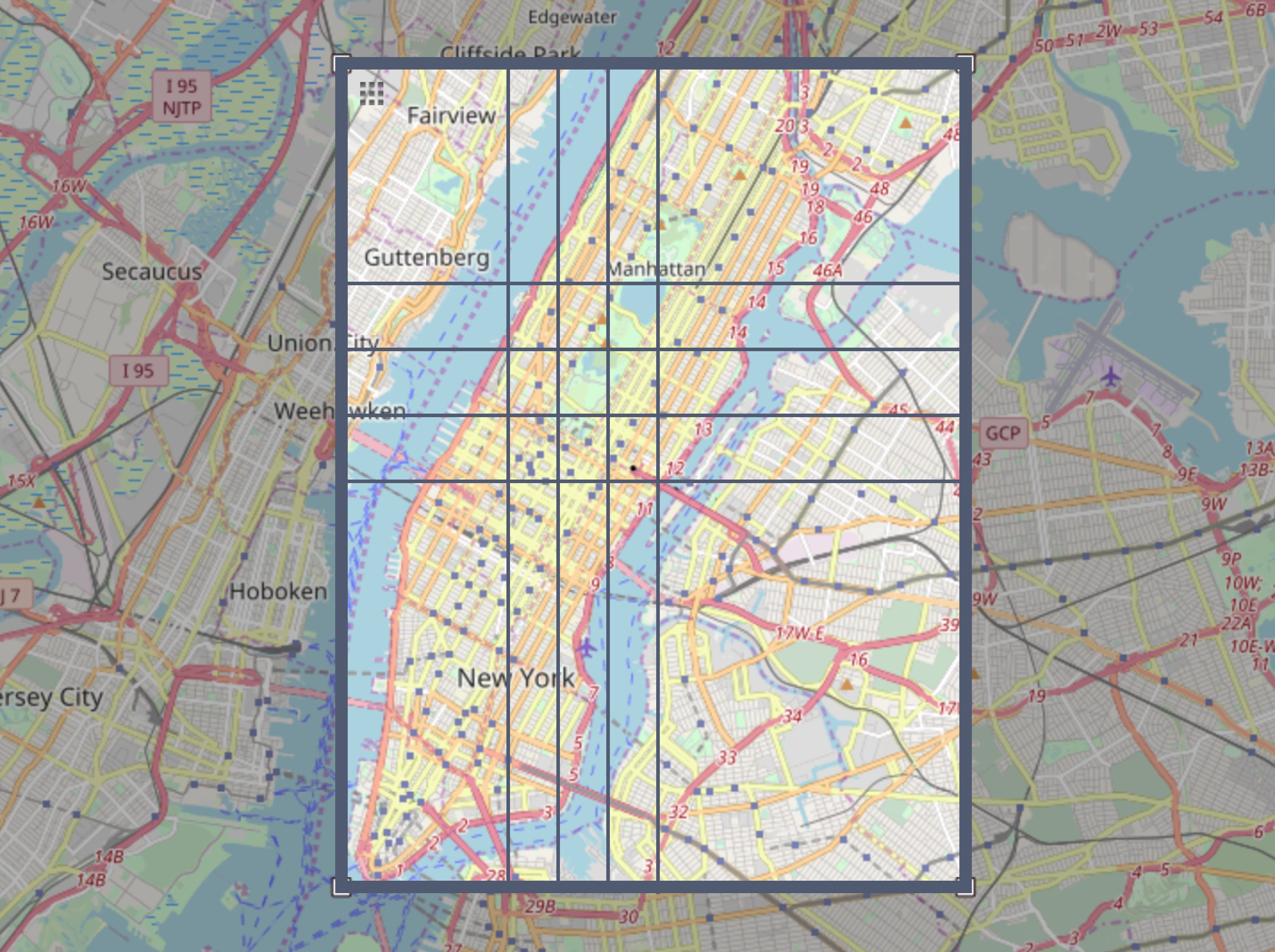}
\caption{Map of the study area in NYC and illustration of the dimensions of the 12x12 grid used.}
\label{fig:map_nyc}
\end{figure}

Having studied the behavior of the proposed DeepJMQR approach and some of its properties with an illustrative example, we now evaluate its performance in a large-scale dataset for a significantly more complex problem and in comparison with various recent state-of-the-art alternatives. To that end, we shall consider a dataset consisting of 1.1 billion taxi trips from New York that were made publicly available by the NYC Taxi \& Limousine Commission \cite{nyctaxidata}. For the purpose of this experiment, we focused only in the area around Manhattan between January 2016 and June 2016. We used the GPS coordinates to discretize the area in a 12x12 grid, where each grid cell has sides of 0.01 decimal degrees, as shown in Figure~\ref{fig:map_nyc}. Similarly, time is discretized by considering 30 minute intervals. The goal is then to make 1-hour-ahead forecasts for the number of taxi pickups in all grid cells, so that taxi operators can optimize their fleet's positioning according to the expected demand in a timely way. The dataset was split into 3 months for training, 1 month of validation data for tuning the hyper-parameters of the different models considered, and 2 months for testing. 

For this particular problem, the proposed DeepJMQR network is using 100 3x3 convolutional filters for the ConvLSTM, dropout and batch normalization \cite{ioffe2015batch}. We compare DeepJMQR with the same baselines used for the motorcycle dataset - linear regression/linear quantile regression models (``Linear QR"), and independent models for the conditional expectation and various quantiles (``Indep. DL") - and also other recent alternative methods from the state of the art. Namely, we consider the joint quantile regression in vector-valued Reproducing Kernel Hilbert Space (RKHS) approach by Sangnier et al. \cite{sangnier2016joint}, which uses constraints in a matrix-valued kernel to address the phenomenon of quantile crossing. We refer to these baselines as ``JQR (RKHS)" and ``QR (RKHS)", depending on whether a joint model for all quantiles or multiple independent models are considered (i.e., $\gamma=10^{-2}$ and $\gamma=\infty$, respectively, according the original paper \cite{sangnier2016joint}). All the  hyper-parameters of the Gaussian kernel used, as well as the value of $\gamma$, were tuned using a grid-search procedure based on the performance on the validation set, as done by the authors in \cite{sangnier2016joint}. However, due to memory constraints, it was only possible to use a subset of train set corresponding to 3000 observations during our experiments on a machine with 64GB of RAM. 

\begin{table*}[!t]
\renewcommand{\arraystretch}{1.1}
\caption{Error statistics and losses for NYC Taxi dataset.}
\label{table:res_loss_nyc}
\centering
\begin{tabular}{l | l l | l l l}
Method & MAE & RMSE & Tilted Loss & Crossing Loss & Num. Crosses \\
\hline
Linear QR & 6.164 ($\pm$ 0.000) & 8.800 ($\pm$ 0.000) &  3078.9 ($\pm$ 0.0) & 1.58 ($\pm$ 0.00) & 35840 ($\pm$ 0) \\
QR (RKHS) & - & - &  3044.4 ($\pm$ 0.3) & 0.12 ($\pm$ 0.01) & 30662 ($\pm$ 1389) \\
JQR (RKHS) & - & - &  3062.0 ($\pm$ 0.2) & 0.01 ($\pm$ 0.00) & 128453 ($\pm$ 3874) \\
MCdropout (Zhu Opt) & 5.967 ($\pm$ 0.045) & 8.597 ($\pm$ 0.053) & 3025.7 ($\pm$ 17.5) & 0.00 ($\pm$ 0.00) & 0 ($\pm$ 0) \\
MCdropout (Gal Opt) & 5.967 ($\pm$ 0.045) & 8.597 ($\pm$ 0.053) & 3018.7 ($\pm$ 13.9) & 0.00 ($\pm$ 0.00) & 0 ($\pm$ 0) \\
Indep. DL & 5.962 ($\pm$ 0.045) & 8.580 ($\pm$ 0.056) & 2852.1 ($\pm$ 3.9) & 170.29 ($\pm$ 19.83) & 292022 ($\pm$ 21663) \\
DeepJMQR & 5.912 ($\pm$ 0.026) & 8.528 ($\pm$ 0.037) & 2857.3 ($\pm$ 10.9) & 0.00 ($\pm$ 0.00) & 408 ($\pm$ 411) \\
\end{tabular}
\end{table*}

\begin{table*}[!t]
\renewcommand{\arraystretch}{1.1}
\caption{Statistics of the prediction intervals for the NYC Taxi dataset.}
\label{table:res_quantiles_nyc}
\centering
\begin{tabular}{l | l l | l l}
Method & ICP 90\% & MIL 90\% & ICP 80\% & MIL 80\% \\
\hline
Linear QR &  0.849 ($\pm$ 0.000) & 28.493 ($\pm$ 0.000) & 0.739 ($\pm$ 0.000) & 19.992 ($\pm$ 0.000) \\
QR (RKHS) &  0.882 ($\pm$ 0.003) & 27.213 ($\pm$ 0.045) & 0.751 ($\pm$ 0.001) & 19.187 ($\pm$ 0.011) \\
JQR (RKHS) &  0.876 ($\pm$ 0.002) & 26.606 ($\pm$ 0.021) & 0.762 ($\pm$ 0.001) & 18.825 ($\pm$ 0.006) \\
MCdropout (Zhu Opt) & 0.922 ($\pm$ 0.001) & 27.368 ($\pm$ 0.099) & 0.869 ($\pm$ 0.001) & 21.323 ($\pm$ 0.077) \\
MCdropout (Gal Opt) & 0.919 ($\pm$ 0.001) & 26.741 ($\pm$ 0.100) & 0.865 ($\pm$ 0.001) & 20.835 ($\pm$ 0.078) \\
Indep. DL & 0.900 ($\pm$ 0.005) & 24.968 ($\pm$ 0.420) & 0.806 /+- 0.009) & 18.433 ($\pm$ 0.421) \\
DeepJMQR & 0.899 ($\pm$ 0.006) & 25.621 ($\pm$ 0.642) & 0.803 ($\pm$ 0.009) & 18.330 ($\pm$ 0.467) \\
\end{tabular}
\end{table*}

Besides the baselines mentioned above, we also consider the use of Monte Carlo dropout \cite{gal2016dropout, gal2016uncertainty} as a way of obtaining uncertainty estimates from deep neural networks. As Gal and Ghahramani \cite{gal2016dropout} showed, dropout - a technique that consists on randomly dropping a fraction of the neuron connections during training \cite{srivastava2014dropout} - can be cast as approximate Bayesian inference in deep Gaussian processes, thereby proving that dropout can be an efficient and theoretically-grounded way of obtaining uncertainty estimates from a deep neural network. The idea is simple and consists in using dropout during testing, by performing multiple forward passes in the network with different sets of connections dropped at random in each pass. It is therefore possible to use the various samples $\{\hat{y}_{m,n,t+k}^{(s)}\}_{s=1}^{S_{\mbox{\tiny MC}}}$ from multiple forward passes to approximate the expected value and variance of the predictive distribution $p(y_{m,n,t+k} | \mathcal{Y}_{1:t})$ as:
\begin{align}
\mathbb{\hat{E}}[y_{m,n,t+k}] &= \frac{1}{S_{\mbox{\tiny MC}}} \sum_{s=1}^{S_{\mbox{\tiny MC}}} \hat{y}_{m,n,t+k}^{(s)},\\
\mathbb{\hat{V}}[y_{m,n,t+k}] &= \sigma^2 + \frac{1}{S_{\mbox{\tiny MC}}} \sum_{s=1}^{S_{\mbox{\tiny MC}}} \Big(\hat{y}_{m,n,t+k}^{(s)} - \mathbb{\hat{E}}[y_{m,n,t+k}]\Big)^2,
\end{align}
where $\sigma^2$ denotes the assumed noise level during the data-generating process $y_{m,n,t+k} = f(\mathcal{Y}_{1:t}) + \epsilon$ with $\epsilon \sim \mathcal{N}(0,\sigma^2)$, and $S_{\mbox{\tiny MC}}$ is the number of Monte Carlo samples. Based on these assumptions and estimates for mean $\mathbb{\hat{E}}[y_{m,n,t+k}]$ and variance $\mathbb{\hat{V}}[y_{m,n,t+k}]$, it is possible to construct prediction intervals with any desired precision \cite{zhu2017deep, gal2016uncertainty}. In our experiments, we refer to this approach as ``MCdropout".

Although MCdropout has been show to provide uncertainty estimates that are useful for various tasks, a key challenge still remains that the uncertainty estimates produces are not well calibrated. Aiming at addressing this issue, several authors have proposed different ways of estimating $\sigma^2$ in order to produce well-calibrated and reliable uncertainty estimates. For example, Yarin Gal \cite{gal2016uncertainty} proposes the use of a grid search procedure in order to optimize $\sigma^2$ over the log probability in a validation set. Similarly, Zhu et al. \cite{zhu2017deep} propose estimating $\sigma^2$ based on the residual sum of squares evaluated on an independent validation set: 
\begin{align}
\hat\sigma^2 = \frac{1}{N_{\mbox{\scriptsize val}}} \sum_{i=1}^{N_{\mbox{\scriptsize val}}} (y_i - \hat{y}_i)^2.
\end{align}
We distinguish these approaches by referring to them as ``MCdropout (Gal Opt)" and ``MCdropout (Zhu Opt)", respectively. 

Table~\ref{table:res_loss_nyc} shows the results obtained for the different approaches.\footnote{We note that the vector-valued RKHS approach from \cite{sangnier2016joint} does not provide estimates for the mean, but only for the quantiles.} We can verify that the quantiles produced by the proposed deep learning architecture based on ConvLSTM layers produce the quantiles with the lowest tilted loss on the test set. However, it is possible to observe that, when using independent models for the quantiles (``Indep. DL"), the crossing loss increases dramatically due to extreme number of quantile crossing cases (almost 300.000 on average). On the other hand, the proposed joint quantile approach, DeepJMQR, is able to obtain a similar average value for the tilted loss, while keeping the crossing loss at practically zero, and reducing the number of quantile crossing to just a few hundred cases. Moreover, we can verify that, by jointly modeling the mean and multiple conditional quantiles simultaneously, the proposed DeepJMQR network is able to also achieve a statistically significant reduction in MAE and RMSE (p-values=$1.17 \times 10^{-5}$ and $2.81 \times 10^{-4}$, respectively). Regarding the quantiles obtained by using MCdropout, the obtained results show that although the quantiles never cross (by definition, according to the way that they are obtained, they can never cross), the tilted loss is significantly higher than the one obtained by DeepJMQR. 

\begin{figure*}[!t]
\centering
\includegraphics[width=0.7\linewidth]{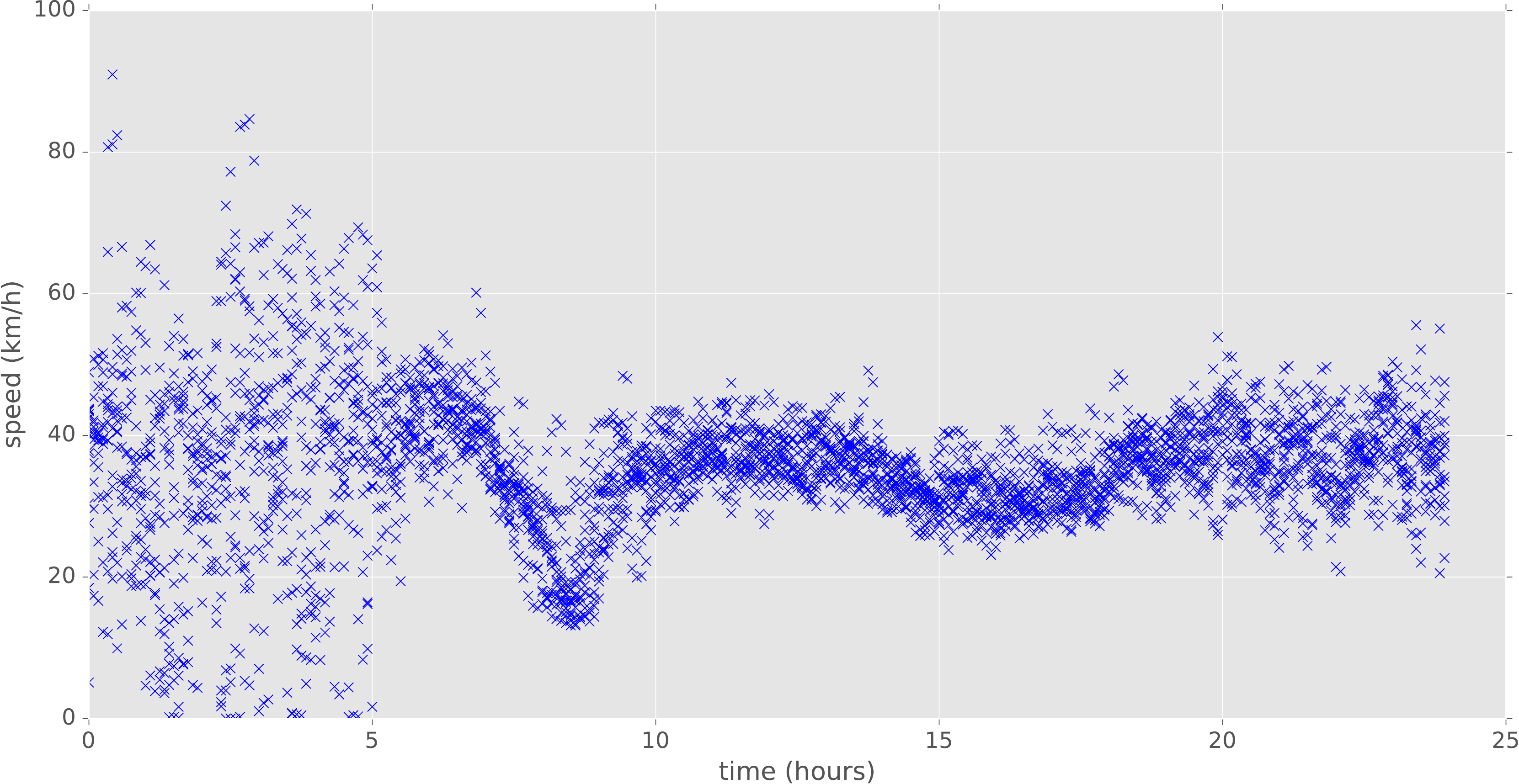}
\caption{Sample of speed observations at a road segment in N{\o}rrecampus for 13 consecutive Thursdays.}
\label{fig:norrecampus_example}
\end{figure*}

In an attempt to better understand the reliability of the quantile estimates produced by the different approaches, we also computed some metrics for evaluating the quality of the 90\% and 80\% prediction intervals. Contrarily to the predictive mean, the quality of the prediction intervals is very difficult to evaluate and quantify, especially using a single criteria. Hence, our main focus lies on the value of the tilted loss in the test set. But, nevertheless, we can try to gain further insights about the prediction intervals produced by the different methods by computing the following statistics:
\begin{itemize}
	\item Interval coverage percentage (ICP), which corresponds to the fraction of the observations that are within the prediction intervals. Hence, for 90\% prediction intervals, this number should be close to $0.90$ or slightly higher;
	\item Mean interval length (MIL), which measures the average length of the prediction intervals. 
\end{itemize}
However, it is important to note that neither of these statistics should be analyzed individually, but rather in the context of each other. For example, it is quite easy to produce prediction intervals with ICP close to $0.90$ but with rather poor MIL. Similarly, it is trivial to obtain prediction bounds with an arbitrarily small MIL, but the ICP would also be lower than desired. 

Having these considerations in mind, Table~\ref{table:res_quantiles_nyc} shows the statistics obtained for the prediction intervals produced by the different methods. As these results show, the majority of the methods are able to provide prediction intervals that contain the expected fractions of observations, since ICP is approximately 0.9 and 0.8 or higher, depending on whether we consider 90\% or 80\% prediction intervals. However, it is possible to verify that the two approaches that use the proposed neural network architecture based on ConvLSTM layers are able to obtain the target coverage percentages while producing narrower intervals. This is a desired property, since tighter bounds with the same coverage percentage are often preferable in practice. For example, in taxi demand forecasting, it means that is possible to ensure that the demand is accommodated with a given level of confidence, while having a lower number of taxis on call in a given area. 

\subsection{Traffic speeds in Copenhagen}

\begin{table*}[!t]
\renewcommand{\arraystretch}{1.1}
\caption{Error statistics and losses for N{\o}rrecampus dataset.}
\label{table:res_loss_norr}
\centering
\begin{tabular}{l | l l | l l l}
Method & MAE & RMSE & Tilted Loss & Crossing Loss & Num. Crosses \\
\hline
Linear QR & 2.183 ($\pm$ 0.000) & 3.950 ($\pm$ 0.000) & 70.25 ($\pm$ 0.00) & 0.33 ($\pm$ 0.00) & 8708 ($\pm$ 0) \\
QR (RKHS) & - & - & 63.92 ($\pm$ 0.11) & 0.04 ($\pm$ 0.00) & 11465 ($\pm$ 466) \\
JQR (RKHS) & - & - & 68.94 ($\pm$ 0.26) & 0.00 ($\pm$ 0.00) & 0 ($\pm$ 0) \\
MCdropout (Zhu Opt) & 1.874 ($\pm$ 0.020) & 3.591 ($\pm$ 0.008) & 74.45 ($\pm$ 0.26) & 0.00 ($\pm$ 0.00) & 0 ($\pm$ 0) \\
MCdropout (Gal Opt) & 1.874 ($\pm$ 0.020) & 3.591 ($\pm$ 0.008) & 68.10 ($\pm$ 0.54) & 0.00 ($\pm$ 0.00) & 0 ($\pm$ 0) \\
Indep. DL & 1.855 ($\pm$ 0.015) & 3.583 ($\pm$ 0.005) & 55.25 ($\pm$ 0.04) & 1.22 ($\pm$ 0.25) & 81810 ($\pm$ 18427) \\
DeepJMQR & 1.817 ($\pm$ 0.007) & 3.557 ($\pm$ 0.004) & 55.33 ($\pm$ 0.10) & 0.00 ($\pm$ 0.00) & 0 ($\pm$ 0) \\
\end{tabular}
\end{table*}

\begin{table*}[!t]
\renewcommand{\arraystretch}{1.1}
\caption{Statistics of the prediction intervals for the N{\o}rrecampus dataset.}
\label{table:res_quantiles_norr}
\centering
\begin{tabular}{l | l l | l l}
Method & ICP 90\% & MIL 90\% & ICP 80\% & MIL 80\% \\
\hline
Linear QR & 0.920 ($\pm$ 0.000) & 10.763 ($\pm$ 0.000) & 0.829 ($\pm$ 0.000) & 6.294 ($\pm$ 0.000) \\
QR (RKHS) & 0.898 ($\pm$ 0.004) & 8.513 ($\pm$ 0.202) & 0.797 ($\pm$ 0.004) & 5.877 ($\pm$ 0.108) \\
JQR (RKHS) & 0.891 ($\pm$ 0.004) & 7.984 ($\pm$ 0.224) & 0.794 ($\pm$ 0.006) & 5.096 ($\pm$ 0.124) \\
MCdropout (Zhu Opt) & 0.959 ($\pm$ 0.001) & 14.011 ($\pm$ 0.025) & 0.936 ($\pm$ 0.001) & 10.917 ($\pm$ 0.019) \\
MCdropout (Gal Opt) & 0.938 ($\pm$ 0.001) & 10.814 ($\pm$ 0.129) & 0.906 ($\pm$ 0.001) & 8.425 ($\pm$ 0.101) \\
Indep. DL & 0.903 ($\pm$ 0.010) & 7.910 ($\pm$ 0.185) & 0.800 ($\pm$ 0.013) & 5.681 ($\pm$ 0.120) \\
DeepJMQR & 0.931 ($\pm$ 0.004) & 8.161 ($\pm$ 0.135) & 0.843 ($\pm$ 0.007) & 5.865 ($\pm$ 0.090) \\
\end{tabular}
\end{table*}

Lastly, the proposed approach was empirically evaluated on a traffic speed forecasting task based on crowdsourced traffic data provided by Google, consisting of 6 months (January 2015 to June 2015) of traffic speeds along 9 consecutive road segments in the N\o rrecampus area in Copenhagen. The latter are part of one of the main accesses into the centre of Copenhagen and are known to be prone to traffic congestion. The dataset was derived from ``Location History" data that Google Maps users agreeingly share with Google. The individual GPS data is aggregated per road segment in 5 minute bins by Google, resulting in a total of 51840 observations per road segment. Kindly notice how this traffic data is very similar to the one that is commonly found at public traffic agencies and local authorities, which is typically obtained by third-party commercial providers such as INRIX and HERE. Although this type of crowdsourced data provides extensive spatial coverage at a lower cost when compared to traditional traffic data collection methods based on expensive road-side equipment, it is know to be susceptible to data quality issues, thus deeming it essential to obtain a fuller picture of the predictive distribution (as the one provided by conditional quantiles) when using it to build forecasting models. For example, in order to ensure that a user reaches a given location on time with a certain level of confidence, it is crucial to rely on an appropriate quantile of the predictive distribution rather than its mean. 

As previously mentioned, due to its nature, crowdsourced traffic data can exhibit high variance at certain periods of the day, especially when the number of probe vehicles (samples) is small. Figure~\ref{fig:norrecampus_example} evidences this phenomenon very clearly, by showing speed observations at an example road segment in N{\o}rrecampus for 13 consecutive Thursdays, thus highlighting the heteroscedastic nature of this dataset, which is particularly evident when contrasting night with day-time periods. 

Similarly to the NYC taxi data, the 6 months of crowdsourced traffic data was split into 3 months for training, 1 month of validation data for tuning the hyper-parameters of the different models considered, and 2 months for testing. However, for this particular problem, the input to the proposed DeepJMQR network at each time step consists of a $9 \times 1$ matrix, and each of two convolutional LSTM layers computes 20 features ($S=20$). Between different layers, dropout is used, with the probability of keeping a connection set to 0.2. 

The proposed approach is compared with same baselines as with the NYC taxi data for the task of forecasting speeds for the next $5$-minute interval. The obtained results are shown in Table~\ref{table:res_loss_norr}. Based on the latter, it is possible to once again verify that the proposed deep learning approach for modeling this spatio-temporal data produces the most accurate quantiles, as evidenced by the fact the obtains by far the lowest values of the tilted loss on the test set. However, we can observe that, while the use of multiple independent models (``Indep. DL") leads to a very significant increase in crossing loss due to the massive average number of quantile crosses, the proposed DeepJMQR completely solves the quantile crossing problem (zero cases of crossing in the test set). Interestingly, as with the NYC data, it seems to be able to do so at the expense of slightly increasing the tilted loss - a similar effect that is observed when comparing the ``QR (RKHS)" and ``JQR (RKHS)" approaches. This suggests that there is a trade-off in place between the quantity of the quantiles (according to the tilted loss) and addressing the quantile crossings problem (crossing loss) by constraining the flexibility of the quantile functions via multi-task learning, which is also acknowledged by \cite{sangnier2016joint}. However, unlike the joint quantile approach in vector-valued RKHS by \cite{sangnier2016joint}, the proposed DeepJMQR network is able to solve the quantile crossing problem, while only increasing the average tilted loss from 55.25 to 55.33.

Regarding the accuracy of the predictions for the mean, it is also possible to again verify that DeepJMQR obtains a significant reduction in MAE and RMSE when compared to all the other approaches (p-values=$2.38 \times 10^{-18}$ and $1.49 \times 10^{-30}$, respectively, when compared to ``Indep. DL"). This is a rather powerful insight. It shows that is possible to obtain better forecasts just by adding additional outputs to a prediction model that correspond to the conditional quantiles, which in practice only incurs on a neglectable increase in training time (an average of 28.19 minutes on a NVIDIA GTX 1080 TI, versus 27.69 minutes for a single model for predicting the mean). Moreover, doing so, provides a much more complete description of the predictive distribution, which is essential for a wide range of practical applications. 

As we did with the experiments with the NYC data, we further computed the ICP and MIL statistics for the prediction intervals obtained by the different approaches for the N\o rrecampus dataset. The obtained results are shown in Table~\ref{table:res_quantiles_norr}. The latter show that the proposed deep learning approaches, together with the approaches based on vector-valued RKHS, obtain the tightest 90\% and 80\% prediction intervals. However, it should be noticed that the ``JQR (RKHS)" approach is able to obtain narrower prediction intervals by not including the the desired fraction of the observations, since the ICP is lower than 0.9 and 0.8, respectively. It is important to note that, when considering for example 90\% prediction intervals, it is often acceptable to obtain ICP values higher than 0.9, but values lower than 0.9 can become problematic. 

\begin{figure*}[!ht]
    \centering
        \subfloat[Linear quantile regression]{\includegraphics[width=1.4\columnwidth]{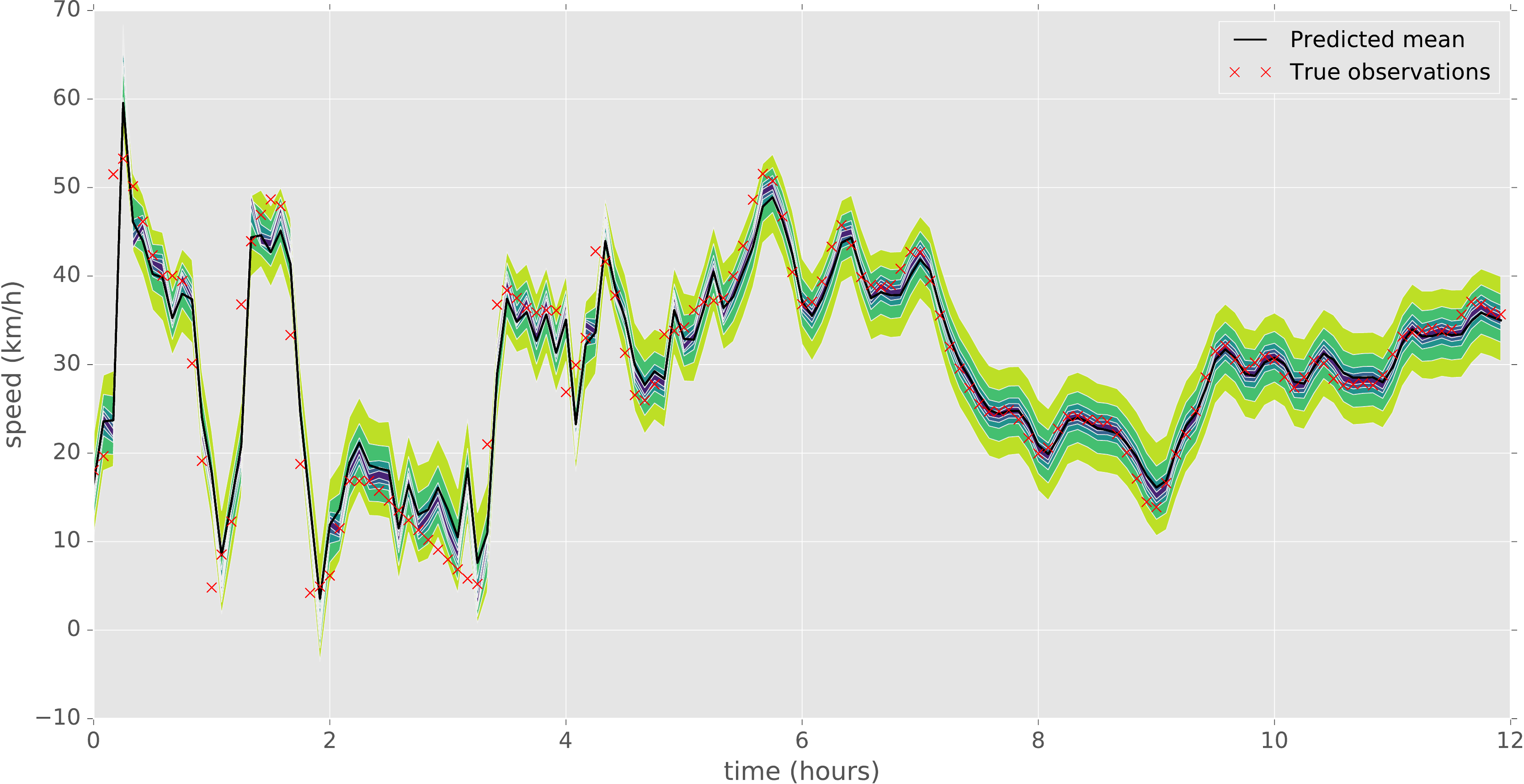}}\\ 
        \subfloat[DeepJMQR]{\includegraphics[width=1.4\columnwidth]{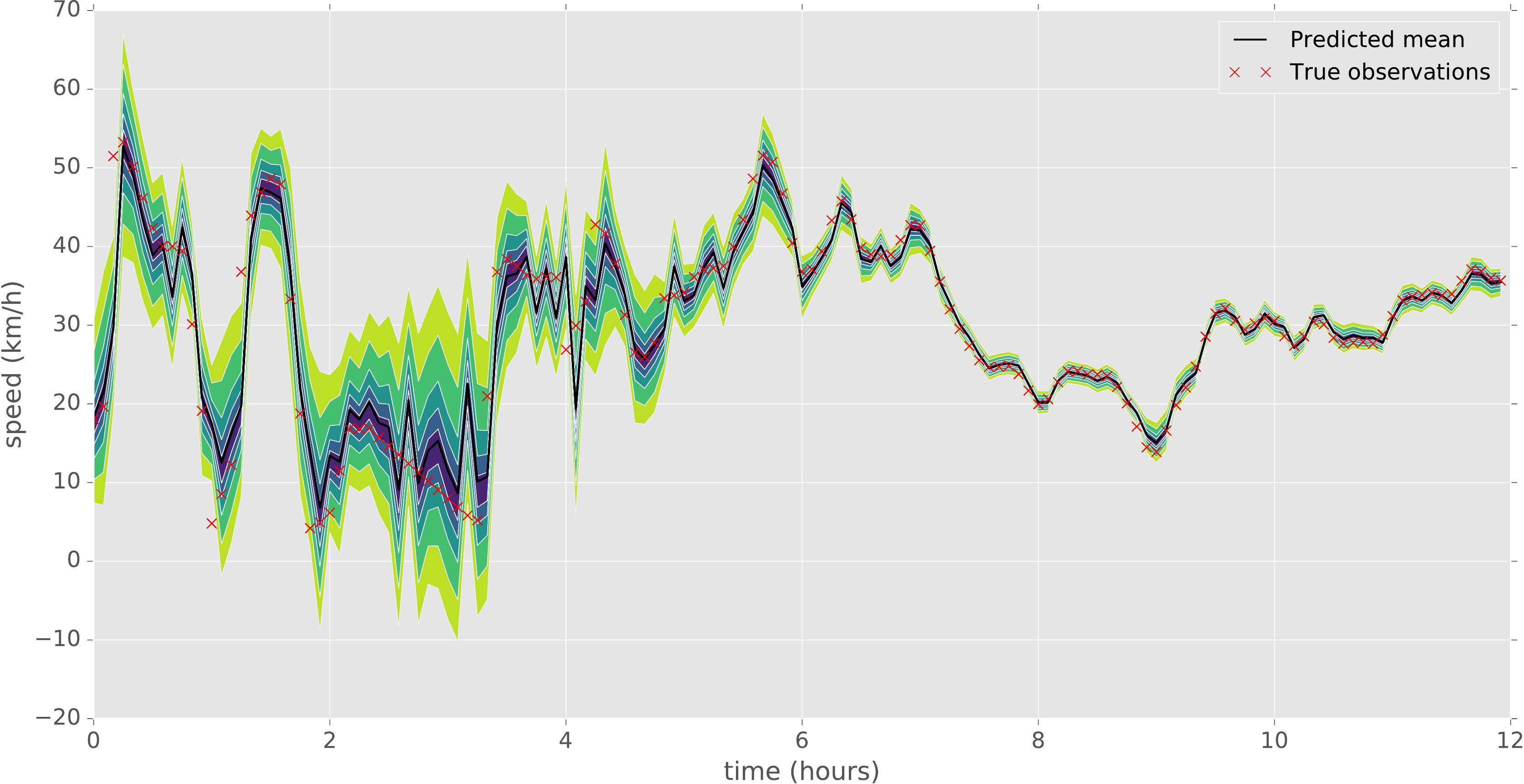}}
\caption{Example of the quantiles predicted by (a) linear quantile regression and (b) DeepJMQR for the traffic speeds in a road segment in N\o rreport.}
\label{fig:norr_quant}
\end{figure*}

Lastly, we would like to point out the significant improvements obtained by DeepJMQR over linear quantile regression (``Linear QR"). In this particular case, the latter corresponds to an autoregressive model - a classical and very popular time-series approach in which the time-series observations at time $t$ are regressed on the previous $L$ observations. However, in this particular case, the assumptions of such an approach can be extremely limiting. Figure~\ref{fig:norr_quant} shows the mean prediction and corresponding quantiles produced by the proposed DeepJMQR network for a random sample of the time-series, in comparison to those obtained by standard linear quantile regression. As the figure demonstrates, the DeepJMQR approach is able to capture the heteroscedastic characteristics of the spatio-temporal data, while the linear methods cannot. This is particularly evident when contrasting the width of the prediction intervals during night and day-time periods. We can observe that the prediction intervals produced by DeepJMQR are significantly narrower during day-time periods, thus expressing the confidence of the model in its predictions during these periods. Kindly notice that such a difference can have very significative consequences for practical applications. In contrast, the intervals during the night period are substantially wider, since the quality of the aggregated speed data is lower as a consequence of the number crowdsourced samples being also lower, which can be also be observed from the higher speed variances as show in Figure~\ref{fig:norrecampus_example}. 

\section{Conclusion}
\label{sec:conclusion}

This article proposed a novel multi-output multi-quantile deep learning approach for jointly modeling several conditional quantiles together with the conditional expectation as a way to provide a more complete ``picture" of the predictive density in spatio-temporal problems, which often can be critical for most practical application. By approaching the problem from a multi-task learning perspective, we showed that is possible to obtain a fuller description of the predictive density that goes beyond the conditional expectation almost for free in terms of computational overhead. In doing so, the proposed DeepJMQR network is able to also address the embarrassing quantile crossing problem due to the consistency and coherence constraints imposed by the hard-parameter sharing between the multiple tasks: mean and multiple quantiles prediction. Moreover, thanks to the additional constraints and information from the conditional quantiles, which we empirically show to induce a regularization effect on the neural network, the proposed deep learning architecture is able to achieve a statistically significant reduction in forecasting error when compared to a similar neural network used just for predicting the conditional expectation. All these contributions were verified empirically through detailed experimentation using a popular heteroscedastic dataset from the literature and two large-scale spatio-temporal datasets from the transportation domain, namely taxi demand prediction and traffic speed forecasting. Through these experiments, we show that the proposed DeepJMQR approach is able to significantly outperform state-of-the-art quantile regression methods as well as other methods for predicting the mean. 

Future work will explore the deployment of the proposed methodology in real-world scenarios by using the predicted quantiles for supply optimization, such as in taxi fleet optimization, dynamic routing and transport system optimization. 

\section*{Acknowledgment}

The research leading to these results has received funding from the People Programme (Marie Curie Actions) of the European Union's Seventh Framework Programme (FP7/2007-2013) under REA grant agreement no. 609405 (COFUNDPostdocDTU), and from the European Union's Horizon 2020 research and innovation programme under the Marie Sklodowska-Curie Individual Fellowship H2020-MSCA-IF-2016, ID number 745673. 

The authors would also like to thank Google for proving access to the N\o rrecampus data used in this work.

\ifCLASSOPTIONcaptionsoff
  \newpage
\fi



\bibliographystyle{IEEEtran}
\bibliography{deep-quantiles.bbl}
%

%

\begin{IEEEbiography}[{\includegraphics[width=1in,height=1.25in,clip,keepaspectratio]{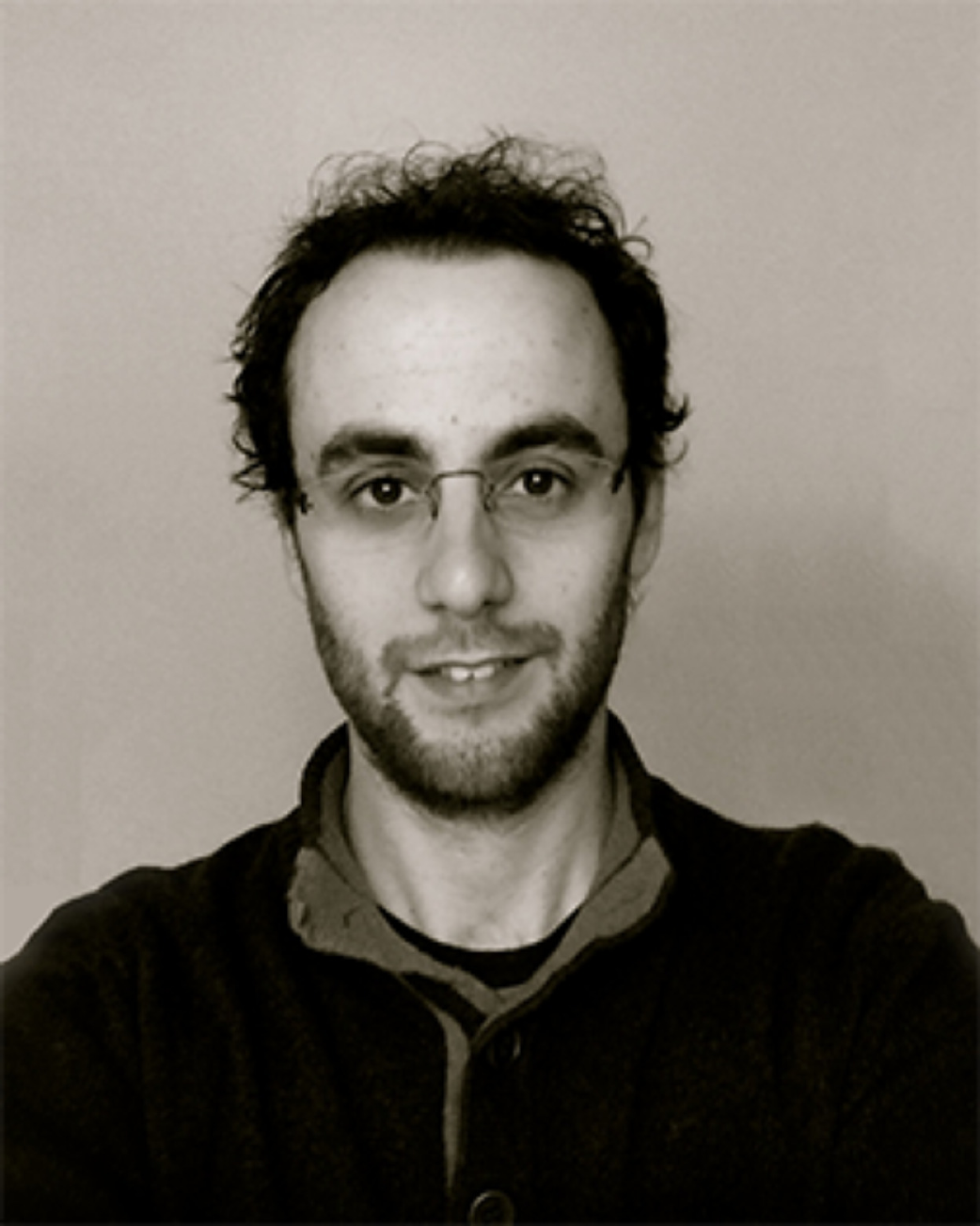}}]{Filipe~Rodrigues}
is Postdoctoral Fellow at Technical University of Denmark, where he is working on machine learning models for understanding urban mobility and the behaviour of crowds, with emphasis on the effect of special events in mobility and transportation systems. He received a Ph.D. degree in Information Science and Technology from University of Coimbra, Portugal, where he developed probabilistic models for learning from crowdsourced and noisy data. His research interests include machine learning, probabilistic graphical models, natural language processing, intelligent transportation systems and urban mobility. 
\end{IEEEbiography}

\begin{IEEEbiography}[{\includegraphics[width=1in,height=1.25in,clip,keepaspectratio]{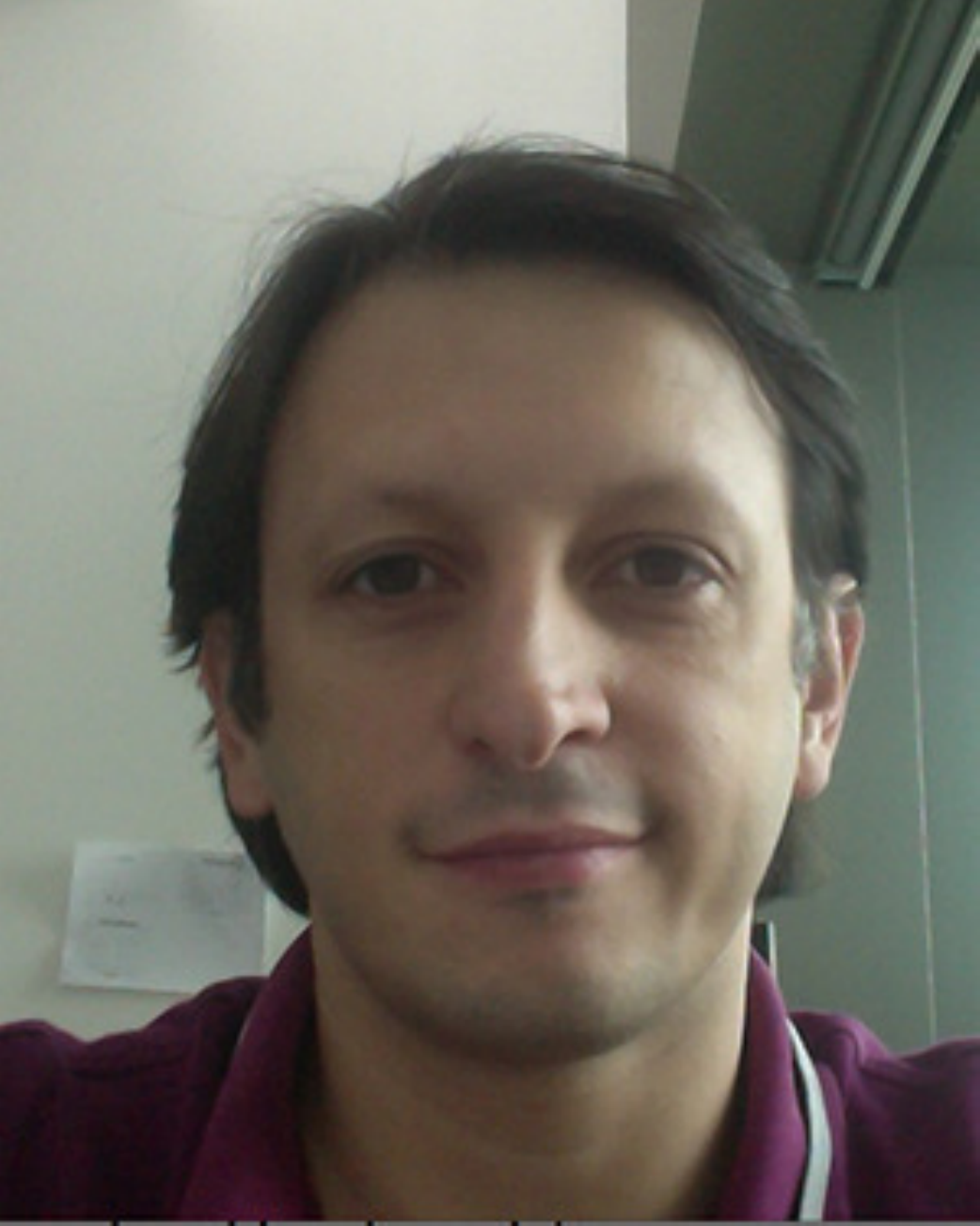}}]{Francisco~C.~Pereira}
is Full Professor at the Technical University of Denmark (DTU), where he leads the Smart Mobility research group. His main research focus is on applying machine learning and pattern recognition to the context of transportation systems with the purpose of understanding and predicting mobility behavior, and modeling and optimizing the transportation system as a whole. He has Masters (2000) and Ph.D. (2005) degrees in Computer Science from University of Coimbra, and has authored/co-authored over 70 journal and conference papers in areas such as pattern recognition, transportation, knowledge based systems and cognitive science. Francisco was previously Research Scientist at MIT and Assistant Professor in University of Coimbra. He was awarded several prestigious prizes, including an IEEE Achievements award, in 2009, the Singapore GYSS Challenge in 2013, and the Pyke Johnson award from Transportation Research Board, in 2015.
\end{IEEEbiography}




\end{document}